\pdfoutput=1

\documentclass[11pt]{article}

\usepackage[final]{acl}

\usepackage{times}
\usepackage{latexsym}

\usepackage[T1]{fontenc}

\usepackage[utf8]{inputenc}

\usepackage{microtype}

\usepackage{inconsolata}

\usepackage{graphicx}

\usepackage{graphicx}

\usepackage{booktabs}       
\usepackage{amsfonts}       
\usepackage{hyperref}       
\usepackage{url}            
\usepackage{xcolor}         

\usepackage{lipsum}
\usepackage{booktabs}
\usepackage{multirow}
\usepackage{graphicx}
\usepackage{tabularx}
\usepackage{subfig}
\usepackage[export]{adjustbox}
\usepackage{amsmath}
\usepackage{enumitem}
\usepackage{float}
\usepackage{placeins}
\usepackage{afterpage}
\usepackage{titlesec}
\usepackage{wrapfig}
\usepackage{pifont}
\usepackage[most]{tcolorbox}
\usepackage{romannum}

\newcommand{\todo}[1]{\textcolor{black}{#1}}

\title{DREsS: Dataset for Rubric-based Essay Scoring on EFL Writing}

\author{Haneul Yoo \hspace{5mm} Jieun Han \hspace{5mm} So-Yeon Ahn \hspace{5mm} Alice Oh \\
    KAIST \\
    \texttt{\{\href{mailto:haneul.yoo@kaist.ac.kr}{\color{black}{haneul.yoo}}, \href{mailto:jieun_han@kaist.ac.kr}{\color{black}{jieun\_han}\href{mailto:ahnsoyeon@kaist.ac.kr}}, \href{mailto:ahnsoyeon@kaist.ac.kr}{\color{black}{ahnsoyeon}}\}@kaist.ac.kr, alice.oh@kaist.edu}
}

\AtBeginDocument{\pagenumbering{arabic}}
\begin{document}
\maketitle

\begin{abstract}

Automated essay scoring (AES) is a useful tool in English as a Foreign Language (EFL) writing education, offering real-time essay scores for students and instructors.
However, previous AES models were trained on essays and scores irrelevant to the practical scenarios of EFL writing education and usually provided a single holistic score due to the lack of appropriate datasets.
In this paper, we release DREsS, a large-scale, standard dataset for rubric-based automated essay scoring with 48.9K samples in total.
DREsS comprises three sub-datasets: $\text{DREsS}_\text{New}$, $\text{DREsS}_\text{Std.}$, and $\text{DREsS}_\text{CASE}$.
We collect $\text{DREsS}_\text{New}$, a real-classroom dataset with 2.3K essays authored by EFL undergraduate students and scored by English education experts.
We also standardize existing rubric-based essay scoring datasets as $\text{DREsS}_\text{Std.}$.
We suggest CASE, a corruption-based augmentation strategy for essays, which generates 40.1K synthetic samples of $\text{DREsS}_\text{CASE}$ and improves the baseline results by 45.44\%.
DREsS will enable further research to provide a more accurate and practical AES system for EFL writing education.\thinspace\footnote{DREsS is publicly available at \url{https://haneul-yoo.github.io/dress/}.}
\end{abstract}

\section{Introduction}
In writing education, automated essay scoring (AES) can provide real-time scores of students' essays to both students and instructors. 
For many students who are hesitant to expose their errors to instructors, the immediate assessment of their essays with AES can create a supportive environment for self-improvement in writing skills~\cite{sun-2022-effects}. For instructors, AES models can ease the time-consuming process of evaluation and serve as a means to validate their assessments, ensuring consistency in their evaluations.

\begin{table*}[tb!]
\centering
\begin{tabular}{@{}ll|rrr@{}}
\toprule 
                                            &           & \textit{Content} & \textit{Organization} & \textit{Language} \\ \midrule
\multicolumn{2}{l|}{$\text{DREsS}_\text{New}$}    & 2,279   & 2,279        & 2,279    \\ \midrule
\multicolumn{1}{l}{\multirow{5}{*}{$\text{DREsS}_\text{Std.}$}} & ASAP P7   & 1,569   & 1,569        & 1,569    \\
                                            & ASAP P8   & 723     & 723          & 723      \\
                                            & ASAP++ P1 & 1,785   & 1,785        & 1,785    \\
                                            & ASAP++ P2 & 1,799   & 1,799        & 1,799    \\
                                            & ICNALE EE & 639     & 639          & 639      \\ \midrule
\multicolumn{2}{l|}{$\text{DREsS}_\text{CASE}$}          & 8,307   & 31,086       & 792      \\ \midrule
\multicolumn{2}{l|}{Total}         & 17,101  & 39,880       & 9,586   \\ \bottomrule
\end{tabular}
\caption{Data statistics of DREsS}
\label{tab:data_stats}
\end{table*}


AES systems can provide either a holistic or an analytic view of essays, but rubric-based, analytical scores are more preferred in the EFL writing education domain~\cite{ghalib-2015-holistic}.
However, there is only a limited amount of rubric-based datasets available for AES, and the rubrics are not consistent in building generalizable AES systems. Furthermore, AES datasets must be annotated by writing education experts because the scoring task requires pedagogical knowledge of English writing. To date, there is a lack of usable datasets for training rubric-based AES models, as existing AES datasets provide only overall scores and/or make use of scores annotated by non-experts. 

\begin{figure}[t]
    \centering
    {\includegraphics[width=\columnwidth]{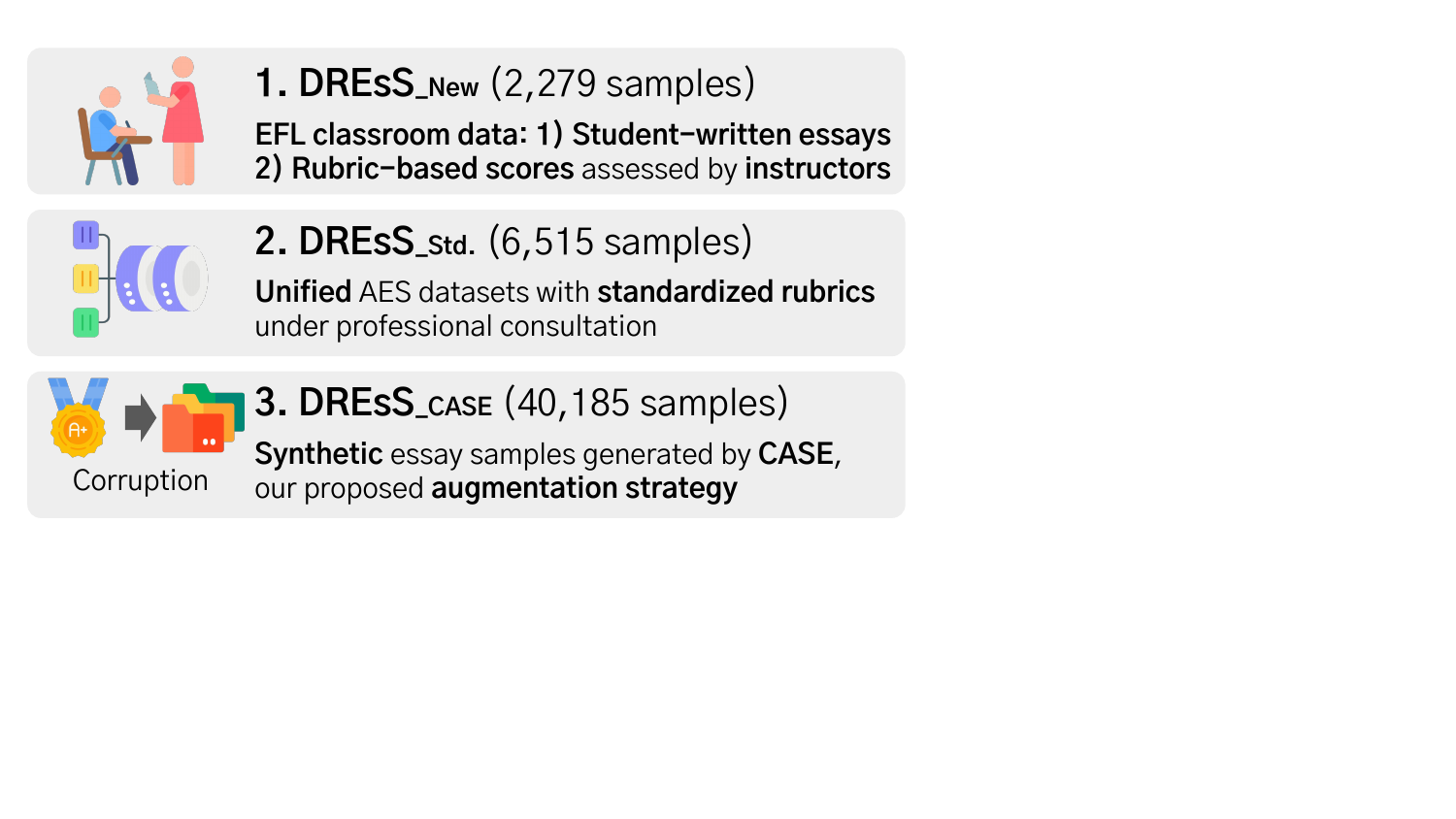}}
    \caption{Data construction of DREsS}
    \label{fig:teaser_image}
\end{figure}

In this paper, we release DREsS (\textbf{D}ataset for \textbf{R}ubric-based \textbf{Es}say \textbf{S}coring on EFL Writing), a large-scale dataset for rubric-based essay scoring using three key rubrics: \textit{content}, \textit{organization}, and \textit{language}.
DREsS consists of three datasets: 1) $\text{DREsS}_\text{New}$ with 2,279 essays from English as a foreign language (EFL) learners and their scores assessed by experts, 2) $\text{DREsS}_\text{Std.}$ with 6,515 essays and scores from existing datasets, and 3) $\text{DREsS}_\text{CASE}$ with 40,185 synthetic essay samples.
We standardize and rescale existing rubric-based datasets to align our rubrics.
We also suggest CASE, a corruption-based augmentation strategy for Essays, employing three rubric-specific strategies to augment the dataset with corruption.
$\text{DREsS}_\text{CASE}$ improves the baseline result by 45.44\%.

\section{Related Work}
In this section, we describe previous studies in automated essay scoring (AES) in terms of the format of predicted scores: holistic AES (\S\ref{sec:holistic_aes}) and rubric-based AES (\S\ref{sec:rubric_based_aes}).
To date, there is only a limited amount of publicly available AES datasets, and their rubrics are inconsistent.
Furthermore, their scores are usually annotated by non-experts lacking pedagogical knowledge in English writing.
Here, we introduce DREsS, a publicly available, large-scale, rubric-based, real-classroom dataset, which can be used as training data for rubric-based AES systems.

\subsection{Holistic AES}
\label{sec:holistic_aes}

\paragraph{ASAP Prompt 1-6}

ASAP dataset\thinspace\footnote{\url{https://www.kaggle.com/c/asap-aes}} is widely used in AES tasks, involving eight different prompts. Six out of eight prompt sets (Prompt 1-6) have a single overall score. 
This holistic AES includes 10K essay scoring data on source-dependent essay (Prompt 3-6) and argumentative essay (Prompt 1-2). 
However, these essays are graded by non-expert annotators, though the essays were written by Grade 7-10 students in the US.

\paragraph{TOEFL11}
TOEFL11~\cite{blanchard-2013-toefl11} corpus from ETS introduced 12K TOEFL iBT essays, which are not publicly accessible now. TOEFL11 only provides a general score for essays in 3 levels (low/mid/high), which is insufficient for building a well-performing AES system.  

\paragraph{Models}
The majority of the previous studies used the ASAP dataset for training and evaluation, aiming to predict the overall score of the essay only~\cite[\emph{inter alia}]{tay-etal-2018-skipflow, cozma-etal-2018-automated, wang-etal-2018-automatic, yang-etal-2020-enhancing}.
Enhanced AI Scoring Engine (EASE)\thinspace\footnote{\url{https://github.com/edx/ease}} is a commonly used, open-sourced AES system based on feature extraction and statistical methods.
In addition, \citet{taghipour-ng-2016-neural} and \citet{xie-etal-2022-automated} released models based on recurrent neural networks and neural pairwise contrastive regression (NPCR) model, respectively.
Still, only a limited number of studies publicly opened their models and codes, highlighting the need for additional publicly available data and further validation of existing models.


\subsection{Rubric-based AES}
\label{sec:rubric_based_aes}

\paragraph{ASAP Prompt 7-8}
ASAP includes only two prompts (Prompt 7-8) that are rubric-based. These two rubric-based prompts consist of 1,569 and 723 essays for each respective prompt. 
The two prompt sets even have distinct rubrics and score ranges, which poses a challenge in leveraging both datasets for training rubric-based models. These essays (Prompt 7-8) are also evaluated by non-expert annotators, similar to ASAP Prompt 1-6.

\paragraph{ASAP++}
To overcome the holistic scoring of ASAP Prompt 1-6, \citet{mathias-bhattacharyya-2018-asap} manually annotated rubric-based scores on those essays.
However, most samples in ASAP++ were annotated by a single annotator, who is a non-expert, including non-native speakers of English.
Moreover, each prompt set of ASAP++ has different attributes or rubrics to each other, which need to be more generalizable to fully leverage such dataset for AES model.

\paragraph{ICNALE Edited Essays}
ICNALE Edited Essays (EE) v3.0 \cite{ishikawa-2018-icnale} presents rubric-based essay evaluation scores and fully edited versions of essays written by EFL learners from 10 countries in Asia. Even though the essays are written by EFL learners, the essay is rated and edited only by a single annotator per sample. They have five native English speakers, non-experts in the domain of English writing education in total. In addition, it is not openly accessible and only consists of 639 samples.

\paragraph{Models}
The scarcity of publicly available rubric-based AES datasets poses significant obstacles to the advancement of AES research. There are industry-driven services such as IntelliMetric®~\cite{rudner-etal-2006-intellimetric} and E-rater®~\cite{blanchard-2013-toefl11, attali-and-burstein-2006-erater}, but none of them are accessible to the public. 
\citet{kumar-etal-2022-many} proposed applying a multi-task learning approach in holistic AES with ASAP and ASAP++, using traits as auxiliary tasks. 
Recent studies have followed up their method, introducing multi-traits AES approaches~\cite[\emph{inter alia}]{chen-li-2023-pmaes, do-etal-2023-prompt, do-etal-2024-autoregressive, lee-etal-2024-unleashing}.
Still, they shed light on predicting a holistic score only due to limited data and built eight different fine-tuned models due to unconsolidated rubrics by each essay prompt.
\todo{Previous studies have explored diverse non-English languages, including Chinese~\cite{song-etal-2020-multi, he-etal-2022-automated}, Japanese~\cite{hirao-etal-2020-automated}, and French~\cite{wilkens-etal-2023-tcfle}, while most of them have mimicked and adapted existing state-of-the-art techniques into non-English languages.}
In order to facilitate AES research in the academic community, it is crucial to release a publicly available rubric-based AES dataset and baseline model.

\section{DREsS Dataset}

\begin{figure*}[htb!]
    \centering
    \subfloat{\centering\includegraphics[width=0.55\textwidth]{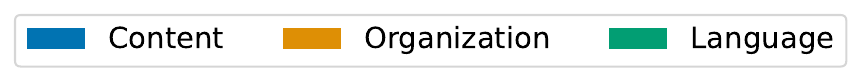}}
    \vspace{-5mm}
    \addtocounter{subfigure}{-1}
    \subfloat[\centering $\text{DREsS}_\text{New}$]{\includegraphics[width=0.4\textwidth]{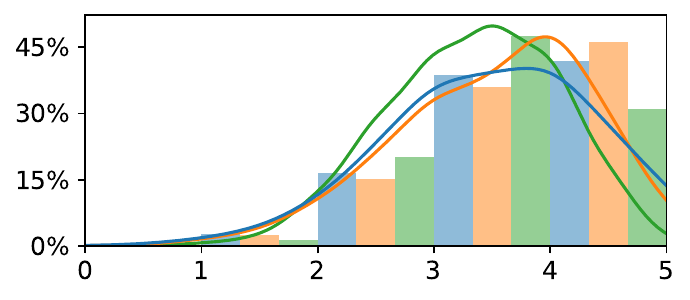}\label{fig:scoure_dist_new}}
    \subfloat[\centering $\text{DREsS}_\text{Std.}$]{\includegraphics[width=0.4\textwidth]{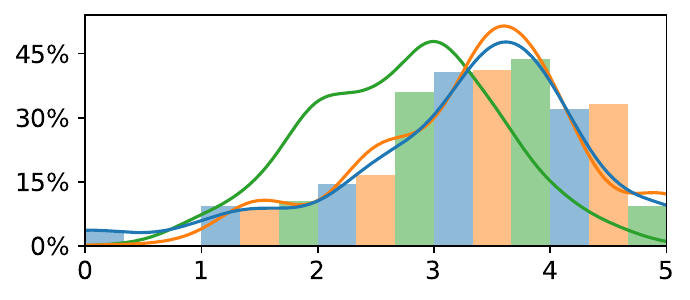}\label{fig:scoure_dist_std}}
    \caption{Score distribution of DREsS}
    \label{fig:score_dist}
\end{figure*}
We construct DREsS with 2.3K samples of our newly collected dataset~(\S\ref{sec:dataset_collection}), 6.5K standardized samples of existing datasets~(\S\ref{sec:standard_data}), and 40.1K synthetic samples augmented using CASE~(\S\ref{sec:synthetic_data_construction}). The detailed number of samples is stated in Table~\ref{tab:data_stats}.


\subsection{Dataset Collection}
\label{sec:dataset_collection}


\begin{table}[htb!]
\begin{tabularx}{\columnwidth}{@{}l|X@{}}
    \toprule
    Rubric & Description \\ \midrule
    \textit{Content} & Paragraph is well-developed and relevant to the argument, supported with strong reasons and examples. \\ \midrule
    \textit{Organization} & The argument is very effectively structured and developed, making it easy for the reader to follow the ideas and understand how the writer is building the argument. Paragraphs use coherence devices effectively while focusing on a single main idea.     \\ \midrule
    \textit{Language} & The writing displays sophisticated control of a wide range of vocabulary and collocations. The essay follows grammar and usage rules throughout the paper. Spelling and punctuation are correct throughout the paper.  \\ \bottomrule
\end{tabularx}
\caption{Rubric explanations}
\label{tab:rubric_explanation}
\end{table}

\paragraph{Dataset Details}
$\text{DREsS}_\text{New}$ includes 2,279 argumentative essays on 22 prompts, having 313.36 words and 21.19 sentences on average. Each sample in DREsS includes students' written essay, essay prompt, rubric-based scores, total score (the sum of three rubric-based scores), and a test type (pre-test, post-test). 
The essays are scored on a range of 1 to 5, with increments of 0.5, based on the three rubrics: \textit{content}, \textit{organization}, and \textit{language}. We chose such three conventional rubrics as standard criteria for scoring EFL essays, according to previous studies from the language education \cite{cumming-1990-expertise, hasan-2022-assessment}. Brief explanations of the rubrics are shown in Table \ref{tab:rubric_explanation}.
The essays are written by undergraduate students whose TOEFL writing score spans from 15 to 21 and enrolled in EFL writing courses at a college in South Korea from 2020 to 2023. 
Most students are Korean and their ages span from 18 to 22, with an average of 19.7.
During the course, students are asked to write an in-class timed essay for 40 minutes both at the start (pre-test) and the end of the semester (post-test) to measure their improvements. 


\paragraph{Annotator Details}
We collect scoring data from 11 instructors, who serve as the teachers of the students who wrote the essays. 
Six of them are non-native speakers, and five of them are native speakers.
All annotators are experts in English education or Linguistics and are qualified to teach EFL writing courses at a college in South Korea.
One instructor was allocated per essay, so the inter-annotator agreement cannot be measured. It follows that an EFL course is usually led by a single instructor, and the essays from the course are assessed by the instructor in a real-classroom setting.
To ensure consistent and reliable scoring across all instructors, they all participate in training sessions with a scoring guide and norming sessions where they develop a consensus on scores using two sample essays. 
Additionally, there was no significant difference among the score distribution of all instructors tested by one-way ANOVA and Tukey HSD at a $p$-value of 0.05.


\subsection{Standardizing the Existing Data}
\label{sec:standard_data}
We standardize and unify three existing rubric-based datasets (ASAP Prompt 7-8, ASAP++ Prompt 1-2, and ICNALE EE) to align with the three rubrics in DREsS: \textit{content}, \textit{organization}, and \textit{language}.
We exclude ASAP++ Prompt 3-6, whose essay type, source-dependent essays, is clearly different from argumentative essays.
We create synthetic label based on a weighted average and then rescale the score of all rubrics into a range of 1 to 5.
Detailed explanations and rationales behind standardizing weights are described in Appendix~\ref{sec:rationale}.
In the process of consolidating the writing assessment criteria, we sought professional consultation from EFL education experts and strategically grouped together those components that evaluate similar aspects under theoretical considerations.

\subsection{Synthetic Data Construction}
\label{sec:synthetic_data_construction}
We construct synthetic data for rubric-based AES to overcome the scarcity of data and provide accurate scores for students and instructors. 
We introduce a corruption-based augmentation strategy for essays (CASE), which starts with a \textit{well-written} essay and incorporates a certain portion of sentence-level errors into the synthetic essay.
In subsequent experiments, we define \textit{well-written} essays as an essay that scored 4.5 or 5.0 out of 5.0 on each criterion.

\vspace{-2mm}
\begin{equation}
    \text{n}(S_c) = \lfloor \text{n}(S_E) * (5.0 - x_i) / 5.0 \rceil
\end{equation}

$\text{n}(S_c)$ is the number of corrupted sentences in the synthetic essay, and $\text{n}(S_E)$ is the number of sentences in the \textit{well-written} essay, which serves as the basis for the synthetic essay.
$x_i$ denotes the score of the synthetic essay.
In this paper, we generate synthetic data with CASE under ablation study for exploring the optimal number of samples.

\paragraph{Content}
We substitute randomly-sampled sentences from \textit{well-written} essays with out-of-domain sentences from different prompts. This is based on an assumption that sentences in \textit{well-written} essays support the given prompt's content, meaning that sentences from the essays on different prompts convey different contents. Therefore, more number of substitutions imply higher levels of corruption in the content of the essay.

\paragraph{Organization}
We swap two randomly-sampled sentences in \textit{well-written} essays and repeat this process based on the synthetic score, supposing that sentences in \textit{well-written} essays are systematically structured in order. The higher number of swaps implies higher levels of corruption in the organization of the essay.

\paragraph{Language}
We substitute randomly-sampled sentences into ungrammatical sentences and repeat this process based on the synthetic score.
We extract 605 ungrammatical sentences from BEA-2019 data for the shared task of grammatical error correction (GEC)~\cite{bryant-etal-2019-bea}. We define ungrammatical sentences with the number of edits of the sentence over 10, which is the 98th percentile. The more substitutions, the more corruption is introduced in the grammar of the essay. We set such a high threshold for ungrammatical sentences because of the limitation of the current GEC dataset that inherent noise may be included, such as erroneous or incomplete correction \cite{rothe-etal-2021-simple}. 

\begin{table*}[htb!]
\centering
\resizebox{\linewidth}{!}{
\begin{tabular}{@{}l|l|ccc|c@{}}
\toprule
\multicolumn{1}{c|}{Model}                                & \multicolumn{1}{c|}{Strategy} & \textit{Content}        & \textit{Organization}   & \textit{Language}       & Total          \\ \midrule
EASE (SVR)                                               & \multicolumn{1}{c|}{\multirow{5}{*}{SFT w/ DREsS}} & -              & -              & -              & 0.360          \\
NPCR~\cite{xie-etal-2022-automated}                      &                         & -              & -              & -              & 0.507          \\ 
\todo{ArTS~\cite{do-etal-2024-autoregressive}}           &                         & \todo{0.601}   & \todo{0.743}   & \todo{\underline{0.592}}   & \todo{\underline{0.690}}           \\ 
BERT~\cite{devlin-etal-2019-bert}                       &                          & \textbf{0.642} & \underline{0.750} & \textbf{0.607} & 0.685 \\
Llama 3.1 8B~\cite{llama3modelcard}                     &                          & \underline{0.631} & \textbf{0.771} & 0.589 & \textbf{0.691} \\ \midrule
\multirow{4}{*}{\texttt{gpt-4o}}               & (A) zero-shot ICL     & 0.310   & 0.322        & 0.231    & 0.304 \\
                                               & (B) five-shot ICL      & 0.361   & 0.475       & 0.367    & 0.428 \\ \cmidrule{2-6}
                                               & (C) rubric explanation   & 0.285   & 0.250       & 0.200    & 0.259 \\
                                               & (D) feedback generation  & 0.313   & 0.268        & 0.230    & 0.290 \\ \bottomrule

\end{tabular}
}
\caption{Baseline results of rubric-based automated essay scoring on DREsS (QWK score)}
\label{tab:aes_result}
\end{table*}

\subsection{Score Distribution}
Figure~\ref{fig:score_dist} shows the score distribution of $\text{DREsS}_\text{New}$ and $\text{DREsS}_\text{Std.}$ ranging from 0 to 5.
The score distribution of the AES dataset shows a left-skewed bell-shaped curve, following the general trends in real-classroom settings.
The scarcity of samples on low scores is because instructors are reluctant to give low scores to increase students' self-efficacy and motivate them to learn~\cite{arsyad-2020-grades}. 
To overcome the imbalance of the dataset, we propose CASE, which can generate synthetic data for all score ranges.
$\text{DREsS}_\text{CASE}$ has the same number of samples per score.

\section{Experimental Result}

\subsection{Baseline Result on DREsS}
Table \ref{tab:aes_result} shows the baseline results of rubric-based AES on DREsS.
We use all three subsplits of DREsS as training data, but $\text{DREsS}_\text{New}$, a subsplit comprising essays and scores from real classroom settings, is used exclusively for the validation and the test sets.
In other words, synthetically unified ($\text{DREsS}_\text{Std.}$) or augmented ($\text{DREsS}_\text{CASE}$) data are reserved for training to avoid incomplete or inaccurate evaluation.
Detailed experimental settings are described in Appendix~\S\ref{sec:experimental_setting}.
We adopt the quadratic weighted kappa (QWK) scores, a conventional metric to evaluate the consistency between the predicted scores and the gold standard scores.

We provide the baseline results on DREsS using holistic AES models from previous studies (\emph{i.e.,} EASE (SVR), NPCR~\cite{xie-etal-2022-automated}, and ArTS~\cite{do-etal-2024-autoregressive}), large language model (\emph{i.e.,} \texttt{gpt-4o} from OpenAI\thinspace\footnote{All following experiments using \texttt{gpt-4o} in this paper was conducted from May 21, 2024 to June 5, 2024 under OpenAI API services.} and Llama 3.1 8B~\cite{llama3modelcard} from Meta), and BERT~\cite{devlin-etal-2019-bert}.
Note that fine-tuned BERT is the model that most state-of-the-art AES systems have leveraged.
We train EASE (SVR), NPCR, ArTS, BERT, and Llama 3.1 with DREsS as supervised fine-tuning (SFT) data.
We also test \texttt{gpt-4o} with four different system prompts as follows: 
\begin{enumerate}[label=(\Alph*)]
    \item in-context learning (ICL) with zero-shot
    \item in-context learning (ICL) with five-shots of writing prompts and essays
    \item asking the model to predict essay scores given detailed rubric explanations
    \item asking the model to predict essay scores and provide essay feedbacks that support their predicted scores.
\end{enumerate}

The detailed prompts are described in Appendix~\ref{sec:aes_chatgpt}.
Considering the substantial length of writing prompt and essay, we were able to provide a maximum of 5 shots for the prompt to \texttt{gpt-4o}.
We divided the samples into five distinct score ranges and computed the average total score for each group. Subsequently, we randomly sampled a single essay in each group, ensuring that its total score corresponded to the calculated mean value.
Asking \texttt{gpt-4o} to score an essay shows high variances among the essays with the same score, implying their limitations to be applied as AES systems.

\subsection{Validation of \texorpdfstring{$\text{DREsS}_\text{Std.}$}{DREsS Std.} and \texorpdfstring{$\text{DREsS}_\text{CASE}$}{DREsS CASE}}

\begin{table*}[htb!]
\resizebox{\textwidth}{!}{%
\centering
\begin{tabular}{@{}l|c|ccc|c@{}}
\toprule
\multicolumn{1}{c|}{Model}                     & Strategy             & \textit{Content} & \textit{Organization} & \textit{Language} & Total \\ \midrule
BERT~\cite{devlin-etal-2019-bert}              & \multirow{5}{*}{SFT w/ $\text{DREsS}_\text{New}$} & \textbf{0.414}   & 0.311        & \textbf{0.487}    & \underline{0.471} \\
Longformer~\cite{beltagy-etal-2020-longformer} &                      & 0.409   & 0.312        & \underline{0.475}    & 0.463 \\
BigBird~\cite{zaheer-etal-2020-bigbird}        &                      & 0.412   & \underline{0.317}        & 0.473    & 0.469 \\
GPT-NeoX-20B~\cite{black-etal-2022-gpt}        &                      & 0.410   & 0.313        & 0.446    & \textbf{0.475} \\
Llama 3.1 8B~\cite{llama3modelcard}            &                      & \underline{0.413}   & \textbf{0.375}        & 0.426    & 0.466 \\ \midrule
\end{tabular}%
}
\caption{Experimental results of rubric-based AES with different LMs using $\text{DREsS}_\text{New}$}
\label{tab:different_lm}
\end{table*}
\begin{table*}[htb!]
\centering
\resizebox{\textwidth}{!}{
\begin{tabular}{@{}l|l|ccc|c@{}}
\toprule
\multicolumn{1}{c|}{Model}                                & \multicolumn{1}{c|}{Strategy} & \textit{Content}        & \textit{Organization}   & \textit{Language}       & Total          \\ \midrule
\multirow{3}{*}{BERT~\cite{devlin-etal-2019-bert}}       & SFT w/ $\text{DREsS}_\text{New}$                        & 0.414          & 0.311          & 0.487          & 0.471          \\
                                                         & {\hspace{2.5mm}+ $\text{DREsS}_\text{Std.}$}  & 0.599          & 0.593          & 0.587          & 0.551          \\
                                                         & {\hspace{5mm}+ $\text{DREsS}_\text{CASE}$}   & \textbf{0.642} & \underline{0.750} & \textbf{0.607} & \underline{0.685} \\ \midrule
\multirow{3}{*}{Llama 3.1 8B~\cite{llama3modelcard}}       & SFT w/ $\text{DREsS}_\text{New}$                        & 0.413          & 0.375          & 0.426          & 0.466          \\
                                                         & {\hspace{2.5mm}+ $\text{DREsS}_\text{Std.}$}  & 0.581          & 0.608          & 0.574          & 0.563          \\
                                                         & {\hspace{5mm}+ $\text{DREsS}_\text{CASE}$}   & \underline{0.631} & \textbf{0.771} & \underline{0.589} & \textbf{0.691} \\ \bottomrule
\end{tabular}
}
\caption{Empirical validation of data expansion in DREsS}
\label{tab:scalable_result}
\end{table*}
Table~\ref{tab:different_lm} shows experimental results of rubric-based AES with different language models.
We train Longformer~\cite{beltagy-etal-2020-longformer} and BigBird~\cite{zaheer-etal-2020-bigbird}, a language model that accepts long input sequences (\emph{i.e.,} 4,096 tokens), considering the substantial length of writing prompts and essays.
In addition, we train GPT-NeoX-20B~\cite{black-etal-2022-gpt} and Llama 3.1 8B, state-of-the-art LLMs.
Nonetheless, exploiting different models does not significantly affect the performance of AES systems.
\citet{xie-etal-2022-automated} also observed that leveraging different foundation models has no significant effect on AES performance, and most state-of-the-art AES methods have still leveraged BERT~\cite{devlin-etal-2019-bert}. 
Therefore, based on these observations, we choose BERT and Llama 3.1 (8B) as a representative model to further evaluate and validate the effectiveness of our dataset, particularly focusing on the benefits of data standardization and synthesis.

We validate the practical benefits of data standardization ($\text{DREsS}_\text{Std.}$) and synthesis ($\text{DREsS}_\text{CASE}$) with empirical results.
Both fine-tuned BERT and Llama 3.1 exhibit scalable results with the expansion of training data (Table \ref{tab:scalable_result}). 
In particular, the model trained with a combination of our approaches outperforms other baseline models by 45.44\%, demonstrating the effectiveness of data unification and augmentation using CASE.
Interestingly, a state-of-the-art LLM (\emph{i.e.,} \texttt{gpt-4o}) does not outperform fine-tuned small-scale language models (\emph{i.e.,} BERT), achieving 0.257 points lower QWK total score.
Existing holistic AES models show their inability to compute rubric-based scores.

\section{Discussion \& Analysis}
\subsection{Ablation Study}
\begin{figure}[tb!]
    \centering
    {\includegraphics[width=\columnwidth]{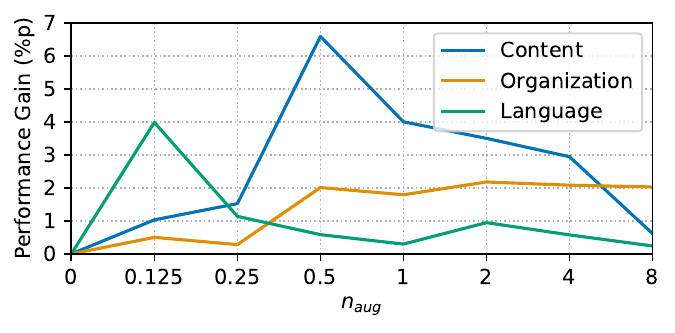}}
    \caption{Ablation experimental results for CASE. $n_{aug}$ is the number of synthetic data by each class per original data among all classes. The x-axis is a log-arithmetic scale.}
    \label{fig:ablation}
\end{figure}
We perform an ablation study to find the optimal number of CASE operations per each rubric.
In Figure~\ref{fig:ablation}, we investigate how the number of CASE operations affects the performance over all rubrics for $n_{aug} = \{0.125, 0.25, 0.5, 1, 2, 4, 8\}$, where $n_{aug}$ denotes the number of synthetic data by each class per original data among all classes (\emph{i.e.,} the ratio of augmented data size compared to the source data size).
CASE on \textit{content}, \textit{organization}, and \textit{language} rubrics show their best performances on 0.5, 2, 0.125 of $n_{aug}$, generating a pair of synthetic essays and corresponding scores in 4.5, 18, 1.125 times, respectively.
We suppose that the detailed augmentation strategies for each rubric and the small size of the original data affect the optimal number of CASE operations.
\textit{Organization}, where corruption was made within the essay and irrelevant to the size of the original data, showed the highest $n_{aug}$.
\textit{Content}, where the corrupted sentences were sampled from 874 \textit{well-written} essays with 21.2 sentences on average, reported higher $n_{aug}$ than language, where the corrupted sentences were sampled from 605 ungrammatical sentences.
Leveraging more error patterns in new grammatical error correction (GEC) data will lead to a scalable increase in the size of $\text{DREsS}_\text{CASE}$ for \textit{language}.

\subsection{CASE vs. Generative Methods}
We verify the quality of synthetic data using CASE compared to generative methods using LLMs.
Here, we use the best-performing baseline rubric-based scoring models trained with DREsS.
We measure a quadratic weighted kappa (QWK) score to measure the similarity between the gold label of the synthetic sample and the predicted score by an AES model.

For LLM to generate synthetic essays, we first give the persona of an EFL student taking an English writing course in a college for students who get TOEFL scores ranging from 15 to 21 and provide five example essays written by EFL students randomly sampled from five distinct score ranges.
We then ask the model to write an essay that matches the rubric-based scores.
The detailed prompts to generate synthetic EFL essays are described in Appendix~\ref{sec:case_vs_generative_methods}.
We randomly sample 900 essays (100 samples per score ranging from 1.0 to 5.0 with an increment of 0.5) from CASE augmentation and synthetic samples generated by \texttt{gpt-4o}.
Table~\ref{tab:case_vs_generative_methods} shows QWK scores of synthetic essays, which validate whether the essays match with their scores.
We use the best-performing baseline rubric-based scoring models in Table~\ref{tab:different_lm}, which only uses $\texttt{DREsS}_\texttt{New}$ as its training and test set.
QWK score of CASE augmentation achieves 0.661 (\emph{substantial agreement}), while the score of the generative method achieves 0.225 on average (\emph{slight to fair agreement}).
Though the detailed persona and example essays are given, \texttt{gpt-4o} fails to write an appropriate level of essays.
Specifically, the predicted rubric-based scores of 900 synthetic essays from \texttt{gpt-4o} across all score ranges are 4.21$_{\pm0.65}$, 4.13$_{\pm0.63}$, and 4.30$_{\pm0.30}$ for \emph{content}, \emph{organization}, and \emph{language}, respectively.

We discuss the benefit of leveraging CASE to generate synthetic essays in EFL writing for three reasons: 1) its difficulty in generating EFL students’ essays, 2) low performance in scoring essays, and 3) controllability and interoperability.
First of all, LLMs are hardly capable of replicating EFL learners’ errors since they are mostly trained with texts from native speakers.
The essays of $\text{DREsS}_\text{New}$ written by EFL students reveal various unique characteristics and error patterns of EFL learners.
Detailed analysis is described in \S~\ref{sec:analysis}.
Second, we found that the state-of-the-art LLM, namely \texttt{gpt-4o}, underperforms in essay scoring tasks compared to BERT-based models, as described in Table~\ref{tab:aes_result}.
Lastly, the black-box nature of LLMs poses challenges in terms of controllability and interpretability.
In contrast, our proposed CASE method offers enhanced control and interpretability.
This mitigates the risks associated with over-reliance on generative methods, fostering a more robust and transparent research approach.

\begin{table}[tb!]
\resizebox{\columnwidth}{!}{%
\centering
\begin{tabular}{@{}l|ccc@{}}
\toprule
                & \textit{Content} & \textit{Organization} & \textit{Language} \\ \midrule
\texttt{gpt-4o} & 0.298            & 0.219                 & 0.158 \\
CASE (Ours)     & 0.625            & 0.722                 & 0.635 \\ \bottomrule
\end{tabular}%
}
\caption{QWK scores of synthetic essays generated by two augmentation methods}
\label{tab:case_vs_generative_methods}
\end{table}

\subsection{In-depth Analysis}
\label{sec:analysis}

\begin{table*}[htb!]
\centering
\begin{tabular}{@{}l|ccc@{}}
\toprule
                                                & DREsS$_\text{New}$  & DREsS$_\text{CASE}$  & \texttt{gpt-4o}     \\ \midrule
\# of sentences *                               & 20.96$_{\pm6.66}$   & 22.67$_{\pm10.10}$   & 16.02$_{\pm2.35}$   \\
\# of tokens *                                  & 313.97$_{\pm96.76}$ & 327.91$_{\pm56.01}$  & 285.84$_{\pm69.07}$ \\
\# of tokens w/o stopwords                      & 162.64$_{\pm49.97}$ & 167.14$_{\pm35.50}$  & 165.49$_{\pm47.91}$ \\
Type-token ratio (TTR) *                        & 0.43$_{\pm0.07}$    & 0.43$_{\pm0.06}$     & 0.51$_{\pm0.04}$    \\
\# of transition signal *                       & 28.03$_{\pm10.3}$   & 28.18$_{\pm14.0}$    & 29.61$_{\pm10.02}$  \\
\# of typos *                                   & 4.39$_{\pm4.11}$    & 6.64$_{\pm8.22}$     & 0.59$_{\pm1.22}$    \\
Flesch reading ease~\cite{flesch-1948-new} *    & 53.66$_{\pm11.84}$  & 59.47$_{\pm11.11}$   & 29.15$_{\pm19.88}$  \\
US grade level~\cite{kincaid-1975-derivation} * & 9.58$_{\pm2.11}$    & 9.12$_{\pm2.58}$     & 13.64$_{\pm3.5}$    \\ \bottomrule
\end{tabular}
\caption{Quantitative analysis. The asterisk denotes a statistically significant difference between \texttt{gpt-4o} and others tested under one-way ANOVA and Tukey HSD test.}
\label{tab:quanti_analysis}
\end{table*}

\begin{table*}[htb!]
\begin{tabularx}{\textwidth}{@{}X|X@{}}
\toprule
DREsS$_\text{New}$                                                                                                                                                                                                                                                                                                                                                                                                                                                                                                 & \texttt{gpt-4o}                                                                                                                                                                                                                                                                                                                                                                                                                                                                                                                                                                                                                                                                                                                                                                                                                                                                                                                                                                     \\ \midrule
\begin{tabular}{@{}p{0.48\textwidth}@{}}In Korea, most of high school students stay with very regular schedule and eat healthy lunch. However, in college they start drinking and often eat fast food. \textcolor{red}{Their health going bad!}\\ College students have much free time and they can do their hobby more. Then, spending time to exercise. Exercise can help \textcolor{red}{college student}. Many college people want to \textcolor{red}{make girl or boyfriend}. If they start exercise and make healthy body, \textcolor{red}{it can be positive effect to people}.\end{tabular} & \begin{tabular}{@{}p{0.48\textwidth}@{}}If I have the chance to visit a foreign country for two weeks, I would like to go to Japan. \textcolor{blue}{Japan is a country that I have always wanted to visit because of its unique culture and beautiful landscapes.} 
 
 \textcolor{blue}{One of the main reasons} I want to visit Japan is because of its culture. Japan has a very different culture from my country, and I am very interested in learning about it. I want to see \textcolor{orange}{the traditional Japanese ceremonies and try wearing a kimono.} I also want to learn about \textcolor{orange}{Japanese tea ceremonies} because they are very famous.
 
 \textcolor{blue}{Another reason} I want to visit Japan is because of its beautiful landscapes. Japan has many \textcolor{orange}{beautiful mountains and forests} that I would like to explore. I have seen pictures of \textcolor{orange}{Mount Fuji}, and I think it is very beautiful. I would like to go hiking there and see the cherry blossoms in spring.
 
 \textcolor{blue}{In conclusion}, I want to visit Japan because it has a unique culture and beautiful landscapes. I think it would be a very interesting and enjoyable trip. \end{tabular} \\ \bottomrule

\end{tabularx}
\caption{Sample essays whose rubric scores on content, organization, and language are all 1 out of 5. Sample-level analysis on \emph{content} (orange), \emph{organization} (blue), and \emph{language} (red).
}
\label{tab:sample_analysis}
\end{table*}

Table~\ref{tab:quanti_analysis} shows quantitative analysis of essays from DREsS$_\text{New}$ and DREsS$_\text{CASE}$ compared to \texttt{gpt-4o} augmentation concerning linguistic features.
Student-written essays in DREsS$_\text{New}$ include unique patterns of \todo{EFL} learners.
For instance, essays in DREsS$_\text{New}$ tend to be longer than synthetic essays from \texttt{gpt-4o}, with more number of sentences but easier and shorter sentences, according to Flesch reading ease~\cite{flesch-1948-new} and the number of tokens, respectively.
Interestingly, EFL students use fewer unique words but frequently use unnecessary stopwords.
Essays from EFL students include typos and spelling errors which cannot be made from the generation outputs of LLMs.
Note that one of the major strengths of the DREsS dataset is the inclusion of errorful essays written by EFL learners in the real-world classroom.

Table~\ref{tab:sample_analysis} shows two sample essays with a score of 1 under the same writing prompt.
The synthetic essay from \texttt{gpt-4o} fails to reflect the EFL learners' errors, generating essays that include \emph{content}, \emph{organization}, and \emph{language} features needed for a well-written essay. For \emph{organization}, the essay from \texttt{gpt-4o} is well-structured with the use of appropriate transition signals and an appropriate thesis sentence in the first paragraph (blue text). For \emph{content}, each body paragraph includes detailed examples to support the argument (orange text). For language, the essay does not include any grammatical errors. In contrast, the essay from DREsS$_\text{New}$ lacks transitional signals, a thesis sentence, and supporting examples. The essay also includes a few grammatical errors and awkward phrases (red text), as it is written by EFL learners in a real-world classroom. 


\section{Conclusion}

We release the DREsS, a large-scale, standard rubric-based essay scoring dataset with three subsets: $\text{DREsS}_\text{New}$, $\text{DREsS}_\text{Std.}$, and $\text{DREsS}_\text{CASE}$.
$\text{DREsS}_\text{New}$ is the first reliable AES dataset with 2.3K samples whose essays are authored by EFL undergraduate students and whose scores are annotated by instructors with expertise.
According to previous studies from language education, we also standardize and unify existing rubric-based AES datasets as $\text{DREsS}_\text{Std.}$.
We finally suggest CASE, corruption-based augmentation strategies for essays, which generates 40.1K synthetic samples and improves the baseline result by 45.44\%.
This work aims to encourage further AES research and practical application in EFL education.

\section*{Limitations}
\label{sec:limitations}
Our research focuses on learning \textit{English} as a foreign language because there already exist datasets, and the current language models perform the best for English. There are many L2 learners of other languages whose writing classes can also benefit from AES. Our findings can illuminate the directions of data collection, annotation, and augmentation for L2 writing education in other languages as well. We leave that as future work.

$\text{DREsS}_\text{New}$ is collected through the EFL writing courses from a college in South Korea, and most of the essays are written by Korean EFL students.
EFL students in different cultural and linguistic backgrounds might exhibit different essay-writing patterns, which might affect the distribution of scores and feedback.
We suggest a further extension of collecting the DREsS dataset from diverse countries.


Our augmentation strategy primarily starts from \textit{well-written} essays and generates erroneous essays along with corresponding scores; therefore, this approach faces challenges in synthesizing \textit{well-written} essays.
However, we believe that \textit{well-written} essays can be reliably produced by LLMs, which have demonstrated strong capabilities in generating high-quality English text.
Also, an optimized rationale (\emph{e.g.,} a threshold in corruption, corruption scale) will advance CASE, which we leave for future work.

We acknowledge that the experimental results in Table~\ref{tab:aes_result}-\ref{tab:different_lm} might not fully cover state-of-the-art models in AES.
Nonetheless, it is noteworthy that those results are a \emph{baseline} for our dataset.
We emphasize that the core contribution of this paper is the construction and the public release of a large-scale AES dataset (DREsS), not a proposal for AES model architecture.
We believe nine different models---namely, state-of-the-art AES-specialized models (EASE, NPCR, ArTS), LLMs (GPT-4o, Llama 3.1, GPT-NeoX), and transformer-based models with different input sizes (BERT, Longformer, BigBird)---sufficiently cover empirical testing of existing models.
We leave examining state-of-the-art AES models for future work, with a proposal of and comparison to a novel architecture.

\section*{Ethics Statement}
\label{sec:ethics_statement}
All studies in this research project were conducted with the approval of our institutional review board (IRB).
Annotators were fairly compensated (approximately USD 18), which exceeds the minimum wage in the Republic of Korea in 2024 (approximately USD 7.3).
To prevent any potential impact on student scores or grades, we requested students to share their essays only after the end of the EFL courses.
We also acknowledged and addressed the potential risk associated with releasing a dataset containing human-written essays, especially considering privacy and personal information.
To mitigate these risks, we plan to 1) employ rule-based coding and 2) conduct thorough human inspections to filter out all sensitive information.
Additionally, access to our data will be granted only to researchers or practitioners who submit a consent form, ensuring responsible and ethical usage.

\section*{Acknowledgment}
This work was supported by Institute for Information \& communications Technology Promotion(IITP) grant funded by the Korea government (MSIP) (No. RS-2024-00443251, Accurate and Safe Multimodal, Multilingual Personalized AI Tutors).
This research project has benefited from the Microsoft Accelerate Foundation Models Research (AFMR) grant program through which leading foundation models hosted by Microsoft Azure along with access to Azure credits were provided to conduct the research.

\bibliography{references/custom, references/anthology_cleaned}

\clearpage
\section*{Appendix}
\appendix
\section{Experimental Settings}
\label{sec:experimental_setting}
\begin{table}[htb!]
\centering
\begin{tabular}{@{}lr@{}}
\toprule
Hyperparameter          & Value  \\ \midrule
Batch Size              & 32     \\
Number of epochs        & 10     \\
Early Stopping Patience & 5      \\
Learning Rate           & 2e-5   \\
Learning Rate Scheduler & Linear \\
Optimizer               & AdamW  \\ \bottomrule
\end{tabular}
\caption{SFT configuration}
\label{tab:model_config}
\end{table}
We split $\text{DREsS}_\text{New}$ into training, validation, and test sets in a 6:2:2 ratio with a random seed of 22.
We use $\text{DREsS}_\text{Std.}$ and $\text{DREsS}_\text{CASE}$, a unified or augmented data as training data only.
Additionally, we separate the training, validation, and test set first and then apply CASE in Table~\ref{tab:aes_result}. In other words, training data does not include augmented essays from high-quality essays in the test set, which prevents data leakage.
The AES experiments except for ArTS, GPT-NeoX-20B, and Llama 3.1 (8B) in Table~\ref{tab:different_lm} were conducted under GeForce RTX 2080 Ti (4 GPUs), 256GiB system memory, and Intel(R) Xeon(R) Silver 4114 CPU @ 2.20GHz (40 CPU cores) with hyperparameters denoted in Table~\ref{tab:model_config}.
Fine-tuning ArTS, GPT-NeoX-20B, and Llama 3.1 (8B) was conducted under Quadro RTX 8000 (4 GPUs), 377GiB system memory, and Intel(R) Xeon(R) Silver 4214R CPU @ 2.40GHz (48 CPU cores) with the same hyperparameters.
LLM inference uses greedy decoding (\emph{i.e.,} temperature 0.0).

\section{LLM Prompting}
This section provides detailed system prompts used for the experiments in this paper.

\subsection{Automated Essay Scoring}
\label{sec:aes_chatgpt}
\begin{table*}[th!]
\begin{tabularx}{\textwidth}{@{}l|X@{}}
\toprule
(A) & {\begin{tabularx}{.9\textwidth}{X}Please score the essay with three rubrics: content, organization, and language.\\ \#\#\# Answer format: \{content: Float, organization: Float, language: Float\}\\Note that the float values of scores are within [1.0, 1.5, 2.0, 2.5, 3.0, 3.5, 4.0, 4.5, 5.0].\\ Please answer only in the above JSON format.\\ \\ \#\#\# prompt: \texttt{<essay prompt>}\\ \#\#\# essay: \texttt{<student's essay>}\end{tabularx}}                                                                                        \\ \midrule
(B) & {\begin{tabularx}{.9\textwidth}{X}Please score the essay with three rubrics: content, organization, and language.\\ \#\#\# Answer format: \{content: Float, organization: Float, language: Float\}\\Note that the float values of scores are within [1.0, 1.5, 2.0, 2.5, 3.0, 3.5, 4.0, 4.5, 5.0].\\ Please answer only in the above JSON format.\\ \\ \textcolor{blue}{\#\#\# Examples 1--5:} \\ \\ \#\#\# prompt: \texttt{<essay prompt>}\\ \#\#\# essay: \texttt{<student's essay>}\end{tabularx}}            \\ \midrule
(C) & {\begin{tabularx}{.9\textwidth}{X}Please score the essay with three rubrics: content, organization, and language.\\ \textcolor{blue}{\texttt{<three rubrics explanation>}}\\ \#\#\# Answer format: \{content: Float, organization: Float, language: Float\}\\Note that the float values of scores are within [1.0, 1.5, 2.0, 2.5, 3.0, 3.5, 4.0, 4.5, 5.0].\\ Please answer only in the above JSON format.\\ \\ \#\#\# prompt: \texttt{<essay prompt>}\\ \#\#\# essay: \texttt{<student's essay}\end{tabularx}}   \\ \midrule
(D) & {\begin{tabularx}{.9\textwidth}{X}Please score the essay with three rubrics: content, organization, and language.\\ \#\#\# Answer format: \{content: Float, organization: Float, language: Float, \textcolor{blue}{content\_feedback: String, organization\_feedback: String, language\_feedback: String}\}\\ Note that the float values of scores are within [1.0, 1.5, 2.0, 2.5, 3.0, 3.5, 4.0, 4.5, 5.0].\\ Please answer only in the above JSON format, \textcolor{blue}{with feedback}.\\ \\ \#\#\# prompt: \texttt{<essay prompt>}\\ \#\#\# essay: \texttt{<student's essay>}\end{tabularx}}                                                                                                                                                                                                                                                                                                                                 \\ \bottomrule
\end{tabularx}
\caption{Four different prompts for \texttt{gpt-4o} to get rubric-based scores in the last four rows of Table~\ref{tab:different_lm}}
\label{tab:aes_prompt_chatgpt}
\end{table*}
Table~\ref{tab:aes_prompt_chatgpt} illustrates four different system prompts used in experiments for Table~\ref{tab:different_lm}.

\subsection{Synthetic Essay Generation}
\label{sec:case_vs_generative_methods}
\begin{tcolorbox}[breakable, enhanced]
        You are an English as a foreign language (EFL) learner taking an English writing course in a college for students who get TOEFL scores ranging from 15 to 21. \\
        \\
        \#\#\# Examples 1--5: \texttt{<five pairs of writing prompts and EFL student's essays>}\\
        \#\#\# Scoring criteria: \texttt{<three rubrics explanation>}\\
        \\
        Write an essay with short paragraphs about the given prompt, of which scores are \texttt{<score>} out of 5.0 for all criteria. Note that the essay should include erroneous patterns or typos from EFL students, according to the score.\\
        \\
        \#\#\# Essay prompt: \texttt{<essay\_prompt>}
\end{tcolorbox}

\section{Rationale Behind Standardizing}
\label{sec:rationale}
The weights are not arbitrarily chosen but were determined through expert consultation and theoretical considerations.
Specifically, ASAP Prompt 7 contains four rubrics---ideas, organization, style, and convention---, while Prompt 8 contains six rubrics---ideas and content, organization, voice, word choice, sentence fluency, and convention. Both sets provide scores ranging from 0 to 3.
For \textit{language}, we first create synthetic labels based on a weighted average. 
This involves assigning a weight of 0.66 to the style and 0.33 to the convention in ASAP Prompt 7, and assigning equal weights to voice, word choice, sentence fluency, and convention in ASAP Prompt 8. 
Stylistic features, such as tone, coherence, and voice, are emphasized as higher-order concerns in writing assessment frameworks, while conventions, such as grammar and punctuation, are considered lower-order concerns. This theoretical understanding, combined with consultation with EFL education experts, informs our decision to assign a higher weight to style, particularly for argumentative essays where persuasive and expressive abilities are crucial~\cite{weigle2002assessing}.
For \textit{content} and \textit{organization}, we utilize the existing data rubric (idea for content, organization as same) in the dataset. 
We repeat the same process with ASAP++ Prompt 1 and 2, which have the same attributes as ASAP Prompt 8.
Similarly, for ICNALE EE dataset, we unify vocabulary, language use, and mechanics as language rubric with a weight of 0.4, 0.5, and 0.1, respectively. 


\section{Additional Experimental Results}

\todo{
\begin{table}[ht!]
\resizebox{\linewidth}{!}{
\begin{tabular}{@{}l|ccc|c@{}}
\toprule
                                            & Content & Organization & Language & Total \\ \midrule
SFT w/ $\text{DREsS}_\text{New}$            & 0.411   & 0.375        & 0.425    & 0.464 \\
\hspace{2.5mm} + $\text{DREsS}_\text{CASE}$ & 0.634   & 0.780        & 0.588    & 0.692 \\
\hspace{2.5mm}+ \texttt{gpt-4o} & 0.452     & 0.377        & 0.408    & 0.467 \\ \bottomrule
\end{tabular}
}
\caption{Experimental results of augmentation techniques in AES models with the identical training steps}
\label{tab:identical_training_step}
\end{table}
We rigorously investigate the efficacy of CASE in training AES models by conducting a more controlled experiment using Llama 3.1 (8B) as a foundation model for supervised fine-tuning (SFT).
We fine-tune the models with $\text{DREsS}_\text{New}$, $\text{DREsS}_\text{CASE}$, and synthetic data generated by \texttt{gpt-4o}.
Differing from the main experiments (Table~\ref{tab:aes_result}--\ref{tab:scalable_result}), Table~\ref{tab:identical_training_step} shows the experimental results where the training steps as 5,000 to ensure that the number of training samples is identical.
Notably, adding synthetic data generated by \texttt{gpt-4o} for fine-tuning shows a minimal impact, especially achieving the worst performance in Language.
Aligning with the findings in \S5.2, generative methods are not applicable for essay augmentations.
}
\section{Datasheet for Dataset}
In this section, we document DREsS following the format of Datasheets for Datasets~\cite{gebru2021datasheets}.
The details on the composition and the collection process of the CSRT dataset are described in the main text.

\subsection{Motivation}
\begin{enumerate}
    \item \textbf{For what purpose was the dataset created?} We aim to construct a large-scale, standard, rubric-based dataset for automated essay scoring (AES) to build AES systems that meet the needs of both instructors and students.
    \item \textbf{Who created the dataset (e.g., which team, research group) and on
behalf of which entity (e.g., company, institution, organization)?} The authors constructed DREsS by 1) collecting new essays and scores from the writing courses in their institution, 2) standardizing existing works, and 3) synthesizing new samples.
    \item \textbf{Who funded the creation of the dataset?} See the Acknowledgments and Disclosure of Funding section.
\end{enumerate}

\subsection{Preprocessing/cleaning/labeling}
\begin{enumerate}
    \item \textbf{Was any preprocessing/cleaning/labeling of the data done (e.g., discretization or bucketing, tokenization, part-of-speech tagging, SIFT feature extraction, removal of instances, processing of missing values)?} No. Instead, we conduct rule-based post-processing and human inspection to filter out sensitive information.
    \item \textbf{Was the “raw” data saved in addition to the preprocessed/cleaned/labeled data (e.g., to support unanticipated future uses)?} N/A
    \item \textbf{Is the software that was used to preprocess/clean/label the data available?} N/A
\end{enumerate}

\subsection{Uses}
\begin{enumerate}
    \item \textbf{Has the dataset been used for any tasks already?} No.
    \item \textbf{Is there a repository that links to any or all papers or systems that use the dataset?} N/A
    \item \textbf{What (other) tasks could the dataset be used for?} DREsS can be used as a training and evaluation dataset for automated essay scoring tasks.
\end{enumerate}

\subsection{Distribution}
\begin{enumerate}
    \item \textbf{Will the dataset be distributed to third parties outside of the entity (e.g., company, institution, organization) on behalf of which the dataset was created?} Yes, the dataset is open to the public who submitted a consent form.
    \item \textbf{How will the dataset will be distributed (e.g., tarball on website, API, GitHub)? } The dataset will be distributed through our website.
    \item \textbf{Will the dataset be distributed under a copyright or other intellectual property (IP) license, and/or under applicable terms of use (ToU)?} The dataset will be distributed under the MIT license.
    \item \textbf{Have any third parties imposed IP-based or other restrictions on the data associated with the instances?} No.
    \item \textbf{Do any export controls or other regulatory restrictions apply to the dataset or to individual instances?} No.
\end{enumerate}

\subsection{Maintenance}
\begin{enumerate}
    \item \textbf{Who will be supporting/hosting/maintaining the dataset?} The authors of this paper will maintain DREsS.
    \item \textbf{How can the owner/curator/manager of the dataset be contacted (e.g., email address)?} The owner/curator/manager(s) of the dataset are the authors of this paper. They can be contacted through the emails on the first page of the main text.
    \item \textbf{Is there an erratum?} We will release an erratum at the GitHub repository if errors are found in the future.
    \item \textbf{Will the dataset be updated (e.g., to correct labeling errors, add new instances, delete instances)?} Yes, the dataset will be updated whenever it can be extended to other red-teaming benchmarks. These updates will be posted on the main web page for the dataset.
    \item \textbf{If the dataset relates to people, are there applicable limits on the retention of the data associated with the instances (e.g., were the individuals in question told that their data would be retained for a fixed period of time and then deleted)?} N/A
    \item \textbf{Will older versions of the dataset continue to be supported/hosted/maintained?} Yes.
    \item \textbf{If others want to extend/augment/build on/contribute to the dataset, is there a mechanism for them to do so?} No.
\end{enumerate}
\end{document}